\documentclass[conference]{IEEEtran}
\IEEEoverridecommandlockouts
\usepackage{cite}
\usepackage{booktabs}
\usepackage{amsmath,amssymb,amsfonts}
\usepackage{caption}
\usepackage{graphicx}
\usepackage{textcomp}
\usepackage{xcolor}
\usepackage{multirow}
\usepackage{subcaption}
\usepackage{algorithm}
\usepackage[noend]{algpseudocode}
\usepackage{hyperref}       
\usepackage{url}            
\def\BibTeX{{\rm B\kern-.05em{\sc i\kern-.025em b}\kern-.08em
    T\kern-.1667em\lower.7ex\hbox{E}\kern-.125emX}}

\begin{document}

\title{Beyond Human Vision: The Role of Large Vision Language Models in Microscope Image Analysis}

\author{
\IEEEauthorblockN{Prateek Verma, Minh-Hao Van, Xintao Wu}
\IEEEauthorblockA{
\textit{University of Arkansas}\\
Fayetteville, AR, USA \\
\{prateek, haovan, xintaowu\}@uark.edu}
}

\maketitle

\begin{abstract}
Vision language models (VLMs) have recently emerged and gained the spotlight for their ability to comprehend the dual modality of image and textual data. VLMs such as LLaVA, ChatGPT-4, and Gemini have recently shown impressive performance on tasks such as natural image captioning, visual question answering (VQA), and spatial reasoning. Additionally, a universal segmentation model by Meta AI, Segment Anything Model (SAM) shows unprecedented performance at isolating objects from unforeseen images. Since medical experts, biologists, and materials scientists routinely examine microscopy or medical images in conjunction with textual information in the form of captions, literature, or reports, and draw conclusions of great importance and merit, it is indubitably essential to test the performance of VLMs and foundation models such as SAM, on these images. In this study, we charge ChatGPT, LLaVA, Gemini, and SAM with classification, segmentation, counting, and VQA tasks on a variety of microscopy images. We observe that ChatGPT and Gemini are impressively able to comprehend the visual features in microscopy images, while SAM is quite capable at isolating artefacts in a general sense. However, the performance is not close to that of a domain expert - the models are readily encumbered by the introduction of impurities, defects, artefact overlaps and diversity present in the images.
\end{abstract}



\section{Introduction}
\label{sec:introduction}

The application of large vision language models (VLMs) extends beyond everyday imagery and has significant potential in the realm of scientific imaging encompassing medical, biological, and material sciences. Scientists in these domains routinely examine microscopy images; such images may not tell a lot in isolation, but when combined with (textual) context for the study, experimental description, and the generally available knowledge of the domain, uncover precious scientific insights, details, and guidance for future research. VLMs are capable of reasoning on multimodal input data, and generating unimodal or multimodal responses, such as receiving an image of cells and producing an image of segmented cells. Traditionally left to experts, such tasks are set to benefit greatly from the developments in VLMs - either now or in the near future, they would simply have more textbook knowledge, less biases, more time, newer perspectives, and less fatigue than an average domain expert. While VLMs are still in their nascency and might struggle with drawing accurate conclusions, their general understanding of a vast number of images and libraries of texts cannot be ignored.

ChatGPT \cite{OpenAI2023GPT4TR} is a state-of-the-art language model with impressive human-like performance. The latest version, GPT-4\footnote{GPT-4 was the latest version from OpenAI at the time of writing.}, has demonstrated exceptional abilities in analyzing images and text and producing coherent and creative content. Gemini \cite{team2023gemini} is another attempt to build a visual chat assistant that can understand images, audio, and text. It has demonstrated good reasoning in math and physics. LLaVA \cite{liu2023visual}, an open-source VLM framework combining pre-trained CLIP and LLM, offers opportunities to further explore and build helpful chat assistants. Diving deeper into image analysis, Segment-anything-model (SAM) \cite{kirillov2023segment} is a new foundational model to segment any image efficiently with only one model instance.

It is here that we intend to evaluate where the VLMs shine and where they dull when it comes to simple and routine scientific tasks concerning microscopy images. Through extensive testing, we aim to uncover the strengths and weaknesses of these models in handling the preliminary understanding, intricate detailing, and hierarchical comprehensibility routinely encountered during scientific image analysis. We identify four fundamental and simple, but important tasks - \textit{classification}, \textit{segmentation}, \textit{counting}, and \textit{visual question-answering (VQA)} and assess the performance of some of the popular language and segmentation models - \textit{LLaVA}, \textit{ChatGPT}, \textit{Gemini} and \textit{SAM} - on them. The evaluation inconspicuously highlights the adaptability of these models to different scientific domains, their ability to understand scientific jargon, and their performance in integrating visual and textual information in drawing accurate depictions.

\begin{figure*}
    \centering
    \includegraphics[width=6.35in]{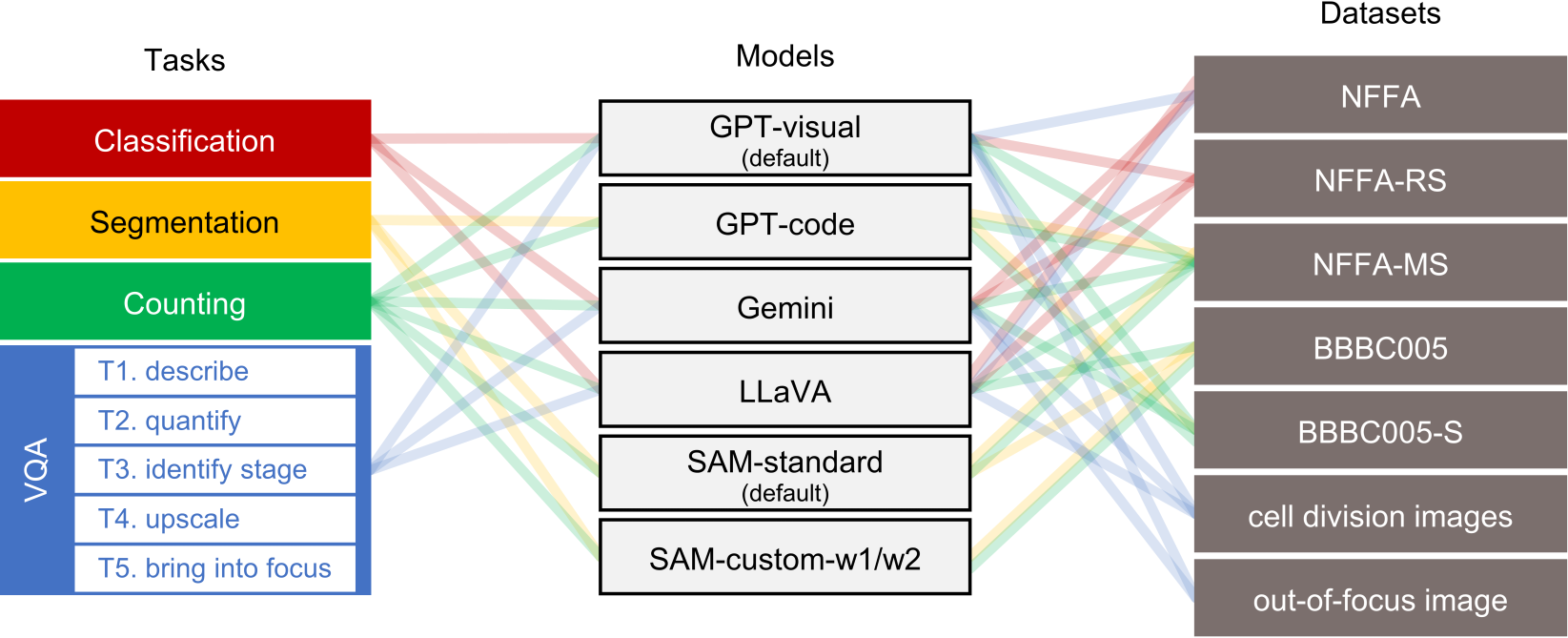}
    \caption{An illustration of the tasks, models, and datasets utilized in this study along with the relational mapping provided as a visual aid to the reader; the details for each box and mapping will become more apparent with their progression in the manuscript.}
    \label{fig:figure_illustration}
\end{figure*}

This paper begins with discussion of some related work followed by the description of the datasets utilized in this study and any subsets created. This is followed by the details of the models used including their capabilities and the different modes in which they were utilized. Undertaken tasks are outlined next, highlighting the levels of complexity and utility to the scientific community. Finally, results and corresponding discussions are presented in the same order as the tasks. Figure \ref{fig:figure_illustration} shows an illustration of the tasks, models, and datasets used in this research.

\section{Related Works}
\label{sec:rel_works}

Prior to language models, materials scientists and biologists resorted to \textit{less recent} machine-learning methods for intelligent processing, examination, and even generation of images. There has been considerable work done in micro-structural identification and classification \cite{shen_generic_2021, ma_image-driven_2020, baskaran_adaptive_2020, tang_machine_2021, tsutsui_methodology_2020}, material property prediction and characterization \cite{perera_optimized_2021, yang_automated_2020, lee_statistical_2020, ilett_application_2020}, process optimization or planning \cite{khatavkar_accelerated_2020, ma_image-driven_2020, yang_automated_2020}, segmentation \cite{akers_rapid_2021, bihani_mudrocknet_2022, chen_deep_2020}, and object identification and quantification \cite{shen_generic_2021, rausch_automated_2022, ilett_application_2020}. Some recent studies have used deep learning for tasks specifically involving microscopy images. A few-shot transfer learning approach for classification of electron backscatter diffraction (EBSD) patterns \cite{kaufmann_efficient_2021} and a semi-supervised few-shot machine learning approach for segmentation of three oxide material systems \cite{akers_rapid_2021} are notable examples.

In recent years, VLMs have caught the attention of the research community due to their impressive abilities in various real-world scientific tasks such as classification, segmentation, object counting, or question-answering on visual inputs. Being trained on a huge amount of data, large models have the capability to execute novel tasks, that are not specified during the training stage. For example, \cite{van2024large} showed that VLMs have impressive performance on biomedical image classification in zero-shot and few-shot settings without the need for retraining. Segment-anything-model (SAM) \cite{kirillov2023segment} was proposed as a foundation model for segmenting any kind of image. To further explore the capability of SAM on medical data, \cite{huang2024segment} adopted 53 open-source datasets including 18 modalities, 84 objects, and more than 1 million 2D images. The study showed that SAM has impressive performance but is unstable and fails in some cases. LLaVA \cite{liu2023visual} introduced a new large multimodal model by connecting CLIP with Llama\cite{touvron2023llama} trained on a visual-instruction tuning dataset. Built upon the success of LLaVA, LLaVA-Med \cite{li2024llava} is a step towards building a multimodal chat assistant for medical data upon previous LLaVA architecture. Being trained on biomedical instruction-following data, the medical chat assistant demonstrated state-of-the-art performance on various tasks. ChatGPT \cite{OpenAI2023GPT4TR}, which is a well-known generative model, has also been evaluated on applications in medical imaging \cite{yan2023multimodal,srivastav2023chatgpt}.

\section{Datasets}
\label{sec:datasets}
Publicly available datasets were selected based on the variety of classes, sample sizes, types of microscopic objects captured, and the availability of appropriate labels and ground truths. 

\begin{figure}
    \centering
    \includegraphics[width=1\linewidth]{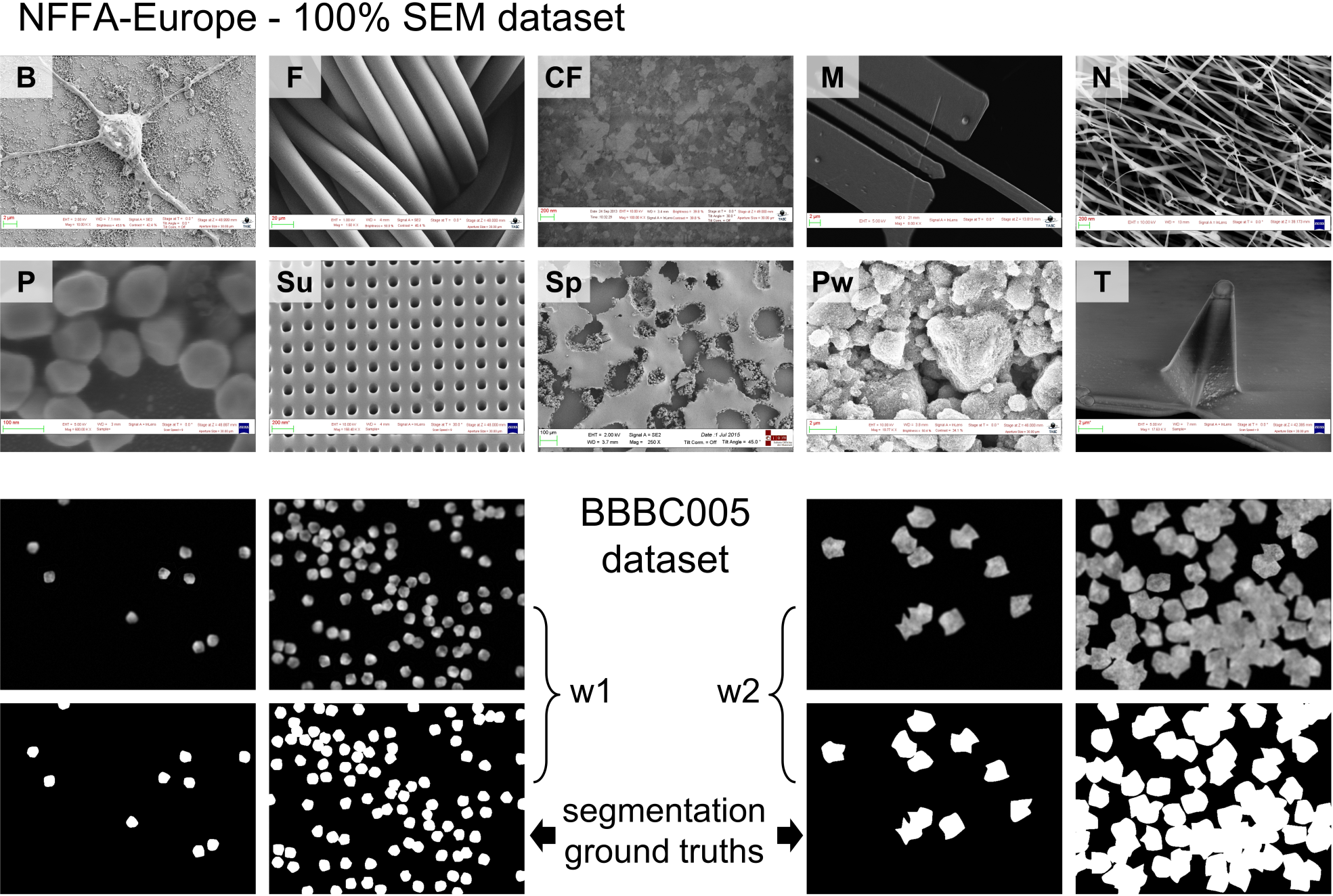}
    \caption{Sample images from each dataset. The classes for the NFFA SEM dataset are defined in Table \ref{tab:datasets_1}. For the BBBC005 dataset, one image each for a low cell count and a high cell count are shown (top row) along with the corresponding segmentation ground truth images below (bottom row). The images on the left belong to the w1 subset, while those on the right belong to the w2 subset.}
    \label{fig:figure_datasets}
\end{figure}

\noindent\textbf{NFFA}: The first dataset that was utilized was the NFFA-Europe - 100\% SEM Dataset (denoted as NFFA henceforth) \cite{aversa_nffa-europe_2018}. This dataset contains 21,169 Scanning Electron Microscopy (SEM) images classified into ten categories such as fibers, particles, MEMS, etc. The images include the SEM instrument label that lists important information such as the electron microscope specifications (such as magnification and beam voltage) used during image capture, scale bar, and instrument manufacturer. Please see Table \ref{tab:datasets_1} for details of classes and the number of images in this dataset. Figure \ref{fig:figure_datasets} (top) shows sample images from the NFFA dataset, exposing the diversity in the kind of visual information available in the SEM images in the ten classes.

As will be evident later, some VLMs and tasks demanded smaller subsets. A 25-image per class, randomly sampled, subset called NFFA-randomly-sampled (or simply NFFA-RS) was created. For certain tasks, a more careful selection of images (by the domain expert among the authors) was desired to preserve the diversity and suitability of the images. A manually selected 25-image subset of the NFFA fibers and NFFA particles classes was thus created. The instrument label was cropped out. This will be referred to as the NFFA-manually-sampled (or simply NFFA-MS). Ground truth label for counting for this subset was manually created by the domain expert.

\begin{table}
    \centering
    \caption{Details of the NFFA-Europe - 100\% SEM Dataset}
    \begin{tabular}{l|l|r}
    \toprule
         Dataset & Classes& Samples\\
         \midrule
         \multirow{10}{*}{NFFA}
             & biological (B)         & 962\\ 
             & fibers (F)             & 150\\
             & coated films (CF)       & 309\\ 
             & MEMS devices (M)      & 4583\\
             & nanowires (N)          & 3815\\ 
             & particles (P)          & 3905\\
             & patterned surface (Su)  & 4752\\ 
             & porous sponge (Sp)      & 174\\
             & powder (Pw)             & 898\\ 
             & tips (T)               & 1621\\
    \bottomrule
    \end{tabular}
    \label{tab:datasets_1}
\end{table}

\noindent\textbf{BBBC005}: The next dataset that was included in this study was from the Broad Bioimage Benchmark Collection (BBBC005) \cite{ljosa_annotated_2012}, which comprises of 19,200 simulated fluorescent cell population images. Of these, half of the images have the cell body stained (w1) and the other half have the nuclei stained (w2). Cell counts were available as ground truth for all images, whereas 1200 images (600 w1 and 600 w2) had segmentation ground truth images available in the form of foreground/background information. Unless otherwise specified, henceforth, BBBC005 will mean the set containing these 1200 images. Figure \ref{fig:figure_datasets} (bottom) shows sample images from the BBBC005 dataset exposing the size difference between the cells in w1 and w2 subsets and the increasing clustering tendency as cell counts increased.

Similar to NFFA, it was required to create smaller sets of the BBBC005 dataset. 50 images each from w1 and w2 were randomly selected and are referred to as BBBC005-sampled (or simply BBBC005-S).

\noindent\textbf{Other data}: Besides these two datasets, a few images sourced from elsewhere were used for certain tasks. These included cell-division images from Dr. Michael W. Davidson's renowned microscopy work at Florida State University \cite{davidson_mitosis_nodate} and an out-of-focus demonstration image from the webpage of the BBBC006 dataset \cite{ljosa_annotated_2012}.

\section{Vision Language Models Used}
\label{sec:vlms}
\noindent\textbf{ChatGPT-4.} OpenAI's GPT-4 has advanced the field of AI with its capacity to process and generate text from combined image and text inputs. As a pre-trained model leveraging Transformer architecture \cite{vaswani2017attention}, GPT-4 has surpassed its predecessor, GPT-3.5, across a majority of scholarly and professional assessments, as well as outperforms other leading models in numerous academic measures. Additionally, it exhibits a reduced tendency to produce factually incorrect information, commonly referred to as `hallucinations' \cite{openaiGPT4TechnicalReport2023}. In the latest revisions, ChatGPT is able to perform computational tasks such as image segmentation, by taking an image input from the user, and writing and executing a Python code, where both the code, and the code output (such as the segmented image) are made available to the user. In this work, GPT-visual refers to the counts obtained from ChatGPT-4 when it was specifically instructed to answer based on the visual cues only and without the use of any computations. GPT-code was the default mode of answer for ChatGPT-4 wherein the model used an image processing method of its choosing until it obtained a valid answer for the counts.

\noindent\textbf{Gemini.} Similar to other language models, Gemini models are trained using advanced model optimization to improve scalability based on Transform-like architecture. The model architecture is not clearly described in \cite{team2023gemini}. However, the visual encoder draws inspiration from Flamingo \cite{alayrac2022flamingo}, which introduced Perceiver Resampler and cross-attention mechanisms to connect visual and linguistic modalities. The models are trained with multimodal inputs including images, audio, and videos interleaving with text from the start. Like ChatGPT, Google Gemini only allows inference via API or web UI. In this study, we collected results using Gemini Pro-Vision 1.0, henceforth referred to as Gemini.

\noindent\textbf{LLaVA.} LLaVA \cite{liu2023visual}, designed as a multimodal visual language assistant, is adept at following instructions that integrate visual and textual elements. It operates as a conversational agent, engaging users with interactions that closely mimic human dialogue. By incorporating a vision encoder, LLaVA gains the capability to interpret images, enabling it to deliver precise responses to a variety of vision-language challenges. This VLM harnesses the strengths of both the pretrained CLIP \cite{radford2021learning} for visual encoding and Llama-2 \cite{touvron2023llama} for linguistic processing, connecting the two with a projection layer that translates visual data into language tokens. These tokens, representing both image and text, are then processed by the language model. As a versatile chat assistant, LLaVA is equipped to handle a series of related questions and answers. Employing a `chain of thought' \cite{wei2022chain} approach, which breaks down complex queries into simpler sub-questions, further augments the model's capabilities.

\noindent\textbf{SAM.} Segment Anything Model \cite{kirillov2023segment}, or SAM (by Meta AI), is a promptable segmentation model that can generate `cut-outs' from any image in accordance with a point, box, or text prompt provided. SAM claims to be capable of zero-shot generalization to unseen objects and images without the requirement of any additional model training, which is why it has been employed to process microscopy images in this study.
While using SAM, it is possible to let SAM run in the `automatic' mode, with no prompts required from the user. In this mode, a pre-specified grid of points defined in SAM's initialization parameters serve as segmentation prompts. The parameter \verb|points_per_side| defines the grid size in pixels. SAM comes with a standard set of parameters (see Table \ref{tab:sam_parameters}), hereon referred to as SAM-standard. Please refer to SAM's documentation for a complete list of parameters and their functions. For certain tasks, some of these parameters were modified (shown in bold in Table \ref{tab:sam_parameters}) and are referred to as SAM-custom. The parameters \verb|pred_iou_thresh| and \verb|stability_score_thesh| that indicate the model's predicted mask quality and stability respectively, and \verb|box_nms_thresh|, \verb|crop_nms_thresh| and \verb|crop_overlap_ratio| that filter duplicate and overlapping masks were the primary parameters that were customized during experiments.

\begin{table}
    \centering
    \caption{SAM-standard and -custom parameters. Parameter names that were modified during tuning are shown in bold face. Custom parameters tuned and finalized on the w1 and the w2 subset after trial-and-error are both shown (comma separated) where they differed from each other.}
    \begin{tabular}{l|c|c}
    \toprule
         Parameter name & SAM-standard & SAM-custom\\
         & & -w1/w2\\
         \midrule
             points\_per\_side & 32 & 32\\ 
             points\_per\_batch & 64 & 64\\
             \textbf{pred\_iou\_thresh} & 0.88 & 0.98, 0.88\\ 
             \textbf{stability\_score\_thresh} & 0.95 & 0.98, 0.95\\
             stability\_score\_offset & 1.0 & 1.0\\ 
             \textbf{box\_nms\_thresh} & 0.7 & 0.9\\
             crop\_n\_layers & 0 & 0\\ 
             \textbf{crop\_nms\_thresh} & 0.7 & 0.95, 0.8\\
             \textbf{crop\_overlap\_ratio} & 512/1500 & 0.05, 0.3\\ 
             crop\_n\_points\_downscale\_factor & 1 & 1\\
             point\_grids & None & None\\
             min\_mask\_region\_area & 0 & 0\\
             output\_mode & binary\_mask & binary\_mask\\
    \bottomrule
    \end{tabular}
    \label{tab:sam_parameters}
\end{table}

\section{Tasks}
\label{sec:tasks}
To demonstrate the capabilities of VLMs in deciphering microscopy images, a variety of tasks were defined keeping in mind the various aspects of analyses that scientists usually perform on such images. Our literature review revealed that segmentation, measurement, counting (representative of a quantification task), visual question answering (VQA), and classification are the most prevalent operations.

\subsection{Classification}
\label{sec:tasks_classification}
VLMs were tasked with figuring out the class to which the micrograph belonged; for example, observe the ten classes of the NFFA dataset. While this task is not challenging for a scientist, it is an important indicator of the abilities of the VLMs as an emerging technology. It is our understanding that the performance of the VLM on classification task is dependant on its ability to comprehend the overall features of, and the artefacts in, the image, possibly along with any textual information present in the annotation label.

NFFA-RS, instead of the entire NFFA, was used with ChatGPT because of the manual nature of the interaction. The prompting strategy involved asking ChatGPT about SEM images and then how would it go about distinguishing between the ten classes. It was subsequently asked to make its best guess as to which class the image belonged to and provide a brief rationale it used to make its judgement. The prompting strategy and an example response from ChatGPT is shown below.
\begin{quote}\small
    \item\textbf{You}: I have SEM images that may belong to any of the following classes - $<$1 to 10. class names$>$. What are the key identifying features of each of these types of SEM images?
    \item\textbf{ChatGPT}: $<$a long answer describing the characteristic features for each class$>$
    \item\textbf{You}: I have SEM images that may belong to any of these classes - $<$1 to 10. class names$>$. Can you please identify the image type using one of the above classes? Please answer with the class label followed by 2-3 sentences of the rationale used. Also list the filename of the uploaded image in the end. Please ask me to upload images indefinitely.
    \item\textbf{ChatGPT}: Absolutely! Please go ahead and upload an SEM image, and I will identify its class from the list you've provided, along with a brief rationale for the classification. You can upload as many images as you like, one at a time, for classification.
    \item\textbf{You}: $<$uploaded image$>$
    \item\textbf{ChatGPT}: Class: (3) Coated Films\newline
    Rationale: The image shows a layered structure with a distinct contrast between the top layer and the underlying material. The top layer appears to be a film with a relatively smooth texture compared to the porous structure beneath it, which suggests a coating over a substrate with a different texture.\newline
    Filename: $<$example filename$>$\newline
    Please upload another image if you need further classifications.
\end{quote}

For classification using Gemini and LLaVa, the entire dataset could be utilized because their APIs could be used to prompt the model in an automated fashion. For LLaVA, a single prompt containing the information about the ten classes and asking the model to output the class in a single word was used as shown below:
\begin{quote}\small
    \item\textbf{You}: $<$uploaded image$>$ I have SEM images that may belong to any of these classes - $<$1 to 10. class names$>$. Can you please identify the image type using one of the above classes? Please answer with the class number only. I want you to make a guess even if you are not sure. I do not want any extra information, only the class number. If it is impossible to determine the class, please answer with `NaN', and nothing else.
    \item\textbf{LLaVA}: 3
\end{quote}

For any task, the prompts used with Gemini were the same as either ChatGPT or LLaVA, unless otherwise specified. 

\subsection{Segmentation}
\label{sec:tasks_segmentation}
Segmentation is an important and routine task for materials scientists and biologists that helps them identify regions (like crystalline vs amorphous) or objects of interest (like nuclei in cells) in an image. Additionally, segmentation is a precursor to many other important tasks. For example, segmentation is essential before any counting or measurement task can be performed. For quantitative evaluation of segmentation, the BBBC005 dataset was used with SAM. Automatic SAM prompting was employed using SAM-standard and SAM-custom parameters. For use with ChatGPT and for the tweaking of SAM parameters, BBBC005-S was used. ChatGPT was prompted to perform the segmentation (using any segmentation algorithm it deemed appropriate), and was asked to output the segmented image like so:
\begin{quote}\small
    \item \textbf{You}: I have SEM images of cells that I want you to segment. Segment the image to obtain a binary segmented image $<$optionally specified to use Otsu's method later$>$. Show the binary segmented image. Let me download the binary segmented image - make sure to use the same filename as that of the uploaded image. I do not want any extra information. Keep asking me to upload images continuously.
    \item \textbf{ChatGPT}: Please upload the first SEM image of the cells that you want to segment.
    \item \textbf{You}: $<$uploaded image$>$
    \item \textbf{ChatGPT}: The segmentation of your SEM image has been completed. You can download the binary segmented image from the following link: $<$download link$>$. Please upload the next SEM image of the cells you want to segment, or let me know if there's anything else you need.
\end{quote}

The NFFA-MS dataset was also used to qualitatively examine the segmentation prowess of SAM. Gemini and LLaVA were not found to be capable of performing segmentation task.

\subsection{Counting}
\label{sec:tasks_counting}
In this task, the VLMs were asked to count the number of featured entities in the images, such as cells, fibers, or particles. The performance on this task depends upon the successful identification (akin to segmentation) of the featured entity by the VLM.

The NFFA-MS was utilized for counting tasks for each model, because the ground truths for counts had to be manually created by the domain expert. In contrast, the BBBC005 dataset, that contains the cell count ground truth labels for each image, was either used in its entirety (for use with LLaVA and SAM-custom) or in its subset form (for inquiry to ChatGPT and Gemini). ChatGPT and Gemini were instructed to count fibers/particles/cells in these images visually and (ChatGPT only) computationally. Note that ChatGPT defaulted to computational evaluation for counting task, unless instructed to answer based on visual cues only. We chose not to specify the computational method to ChatGPT for the counting task. The prompting strategy with ChatGPT was:
\begin{quote}\small
    \item \textbf{You}: I want you to count the number of fibers/particles/cells in an SEM image in two distinct ways.\newline
    Task 1: Estimate the number based on visuals only. Format your answer as `Visual count: $<$count$>$'. Do not share any extra information.\newline
    Task2: Use a smoothing and thresholding approach and remove small objects and count the number of fibers. Do not show analysis, but show the processed images. Format your answer as `Computational count: $<$count$>$', `processed image'. Do not share any extra information.\newline
    Continuously ask me to upload images.
    \item \textbf{ChatGPT}: Please upload the SEM image for analysis.
    \item \textbf{You}: $<$uploaded image$>$
    \item \textbf{ChatGPT}: Visual count: 4\newline
    $<$outputs processed image or a panel of images depending on the segmentation algorithm used$>$\newline
    Computational count: 9
\end{quote}

Counting using LLaVA was performed using a single prompt in a zero-shot setting like so:
\begin{quote}\small
    \item\textbf{You}: $<$uploaded image$>$ I am providing a microscope image of fibers/particles/cells. Based on visual observation, please provide a count of the cells in the image in the format `Count: $<$count$>$'. If you are unable to count the cells, please write NaN.
    \item\textbf{LLaVA}: 5
\end{quote}

SAM was also utilized for the counting task: segments were either first created using SAM-standard or SAM-custom and the number of segments were counted. The counting was either done manually - by the domain expert wherein segments belonging to the background or objects not-of-interest were discarded, or automatically - wherein all segments identified by SAM were counted. As expected, manual counting was only possible for the subsets.

\subsection{Visual Question Answering (VQA)}
\label{sec:tasks_vqa}
In this task, the VLMs were qualitatively evaluated on a set of questions on a variety of microscopy images. The questions were constructed in a way that would demonstrate the VLMs utility to laypeople, domain users, students, and researchers. ChatGPT, LLaVA, and Gemini were first asked to \textit{describe} the provided SEM images (VQA-T1). They were then asked a question related to \textit{quantification} of the object of interest - such as estimating the size or the number of the object(s) (VQA-T2). In the third kind of task (VQA-T3), the model was asked to identify the \textit{stage of a process} shown by a single image (in this case, the stage of cell division) after being asked to describe a similar image. Subsequently, the models were asked to \textit{upscale} a particular image of cell division (VQA-T4). Finally, the models were asked to bring an out-of-focus microscope image \textit{into focus} (VQA-T5).


\section{Results and Discussion}
\label{sec:results}

\subsection{Classification}
\label{sec:results_classification}

\begin{figure}
    \centering
    \includegraphics[width=3.2in]{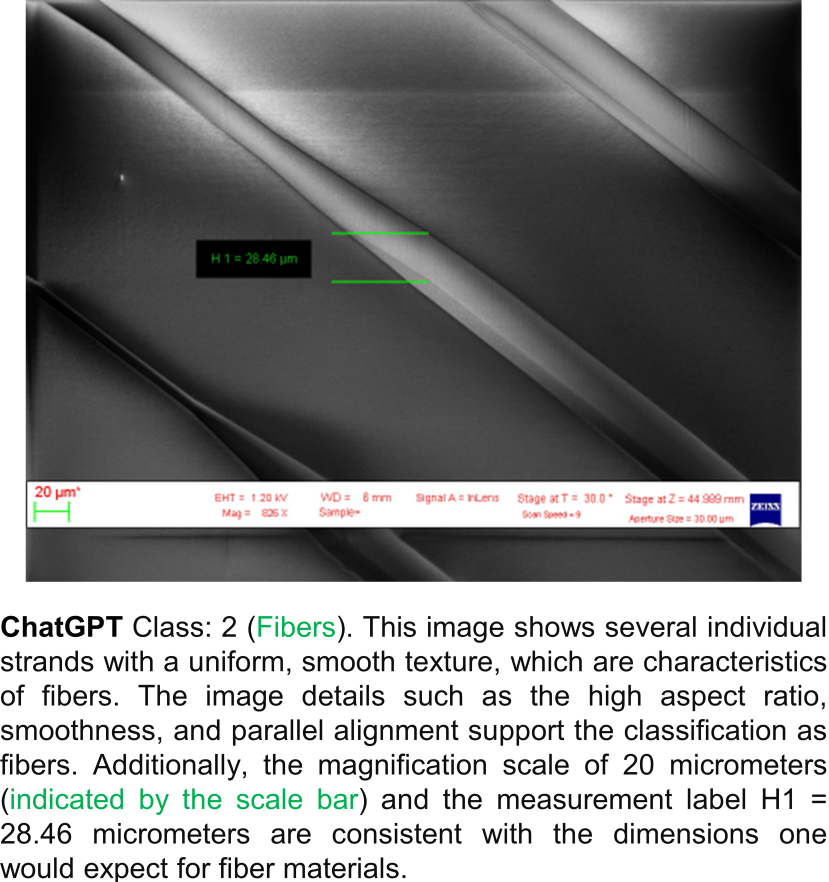}
    \caption{ChatGPT considered the scale bar while predicting the class of this image as fibers correctly. It is possible that this judgement helped ChatGPT to distinguish it from nanowires. Correct key responses are colored in green.}
    \label{fig:figure_chatgpt_scale_bar}
\end{figure}

\begin{figure*}
    \centering
    \includegraphics[width=6.5in]{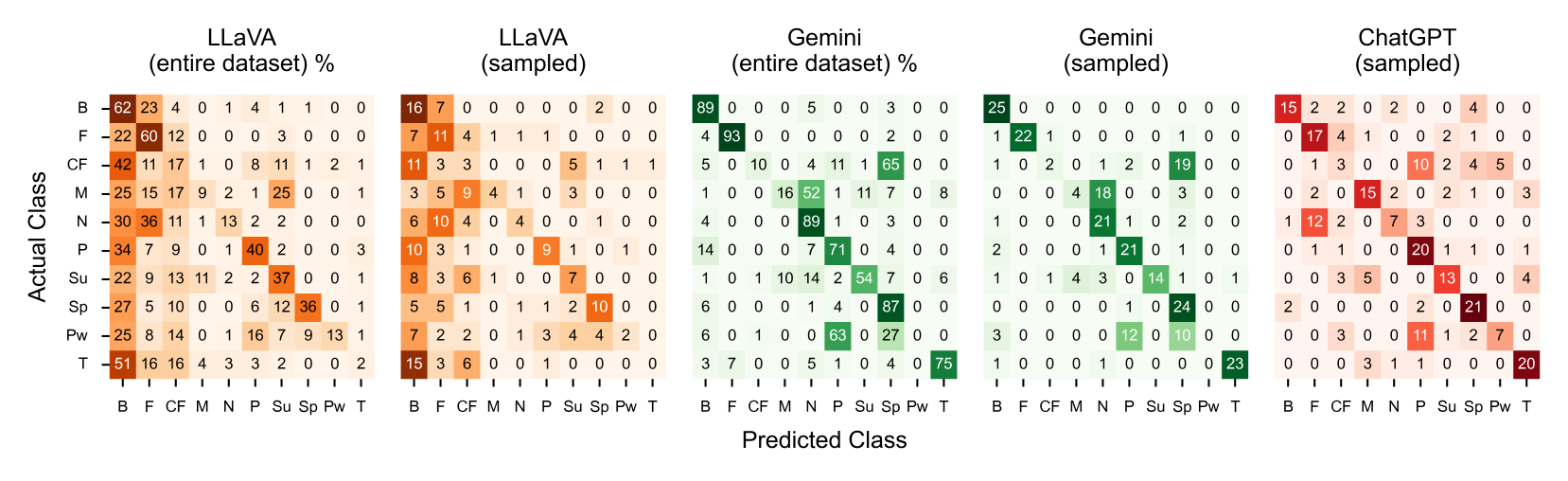}
    \caption{Confusion matrices for the classification of ten categories in the NFFA dataset by (a) LLaVA on the entire dataset (numbers denote percentages with respect to the number of actual samples present in the respective classes), (b) LLaVA on 25 images per class of the NFFA-RS, (c) Gemini on the entire dataset (numbers denote percentages as in (a)), (d) Gemini on 25 images per class of the NFFA-RS, and (e) ChatGPT on 25 images per class of the NFFA-RS. Class names have been abbreviated as noted in Table \ref{tab:datasets_1}.}
    \label{fig:figure_classification}
\end{figure*}

Figure \ref{fig:figure_classification} shows the performance of LLaVA, Gemini, and ChatGPT on the task of distinguishing among 10 categories of SEM images in the NFFA dataset. Overall, ChatGPT did a decent job at understanding the type of image. It was good (exhibited more than 60\% accuracy) at identifying biological samples, fibers, MEMS, particles, sponges, and tips and performed poorly (exhibited less than 28\% accuracy) at identifying powders, nanowires, and coated films. Interestingly, ChatGPT found it difficult to distinguish nanofibers from fibers, revealing that it wasn't automatically considering the scale bar while making this judgement. It must be noted that in several other cases, ChatGPT did, in fact, consider the scale bar, as evident from the rationale it provided (see Figure \ref{fig:figure_chatgpt_scale_bar}). Unsurprisingly, it was also having a difficult time distinguishing among powders, patterned surfaces, and coated films: these are difficult tasks for even a trained scientist largely due to the absence of 3D information and context. More importantly, the rationale provided by ChatGPT was found to be qualitatively quite impressive in almost all cases, even when its class determination was wrong. For example, when provided with a powdered sample, ChatGPT wrongly classified it as `particles', but included this phrase in its rationale: \textit{The particles exhibit a rough texture and irregular shapes, which could be indicative of a powdered substance.}

The confusion matrix for LLaVA is normalized by the actual number of samples in the given class. Note that LLaVA could utilize the entire dataset due to its API. Overall, LLaVA performed worse than ChatGPT and Gemini over the ten classes. Only biological and fiber samples were identified with greater than 60\% accuracy. In the same tone, many samples were wrongly identified as biological, fibers, and coated films, exhibiting the bias of LLaVA in favor of these certain classes; these classes appeared first in the prompt which might have introduced this bias. Similar to ChatGPT, the misclassification by LLaVA was also found to be reasonable to an extent. For example, 37\% nanowires being wrongly classified as fibers, 25\% MEMS being wrongly classified as patterned surfaces, and 17\% powder samples being wrongly classified as particles, is understandable because these classes may visually appear similar to a domain expert as well. The performance of LLaVA on the NFFA-RS is also included for direct comparison with ChatGPT. The results show no considerable difference between the subset and the entire dataset.

Gemini achieved the overall accuracy of 62.40\% and 56.64\% on NFFA-RS and NFFA, respectively, which are significantly better than LLaVA and comparable to ChatGPT. However, the performance of Gemini stood out on its identification of the images of biological samples, fibers, nanowires, porous sponges, and tips, all of which resulted in accuracies greater than 75\%. Similar to ChatGPT, Gemini struggled to identify powders and classified them as either particles (a reasonable error) or sponges. Additionally, unlike ChatGPT or LLaVA, it struggled to identify coated films correctly and consistently classified them as porous sponges. Like LLaVA, there was no considerable difference between the subset and the entire dataset.

LLaVA utilizes pretrained CLIP, which is based on Vision-Transformer architecture, as the visual encoder to process input images. ChatGPT and Gemini have not released the details of their network architectures. According to Gemini's technical report \cite{team2023gemini}, the models were trained with multimodal data from the beginning, and the vision encoder was inspired by the previous work on Flamingo \cite{alayrac2022flamingo}, which is Normalizer-Free ResNet (NFNet). From the explored details, we are intuitively able to understand why Gemini achieved impressive performance in the classification task - the model is multimodally trained from scratch instead of using a pretrained vision encoder.

\subsection{Segmentation}

Quantitative evaluation of the segmentation task was done primarily using dice score for both SAM and ChatGPT. Calculation of dice score was possible for the BBBC005 dataset because the segmented ground truths were available. Additionally, some select segmented images were observed carefully by the domain expert and are presented here with a commentary on the quality of segmentation performed by SAM and ChatGPT.

\begin{figure}
    \centering
    \includegraphics[width=3.2in]{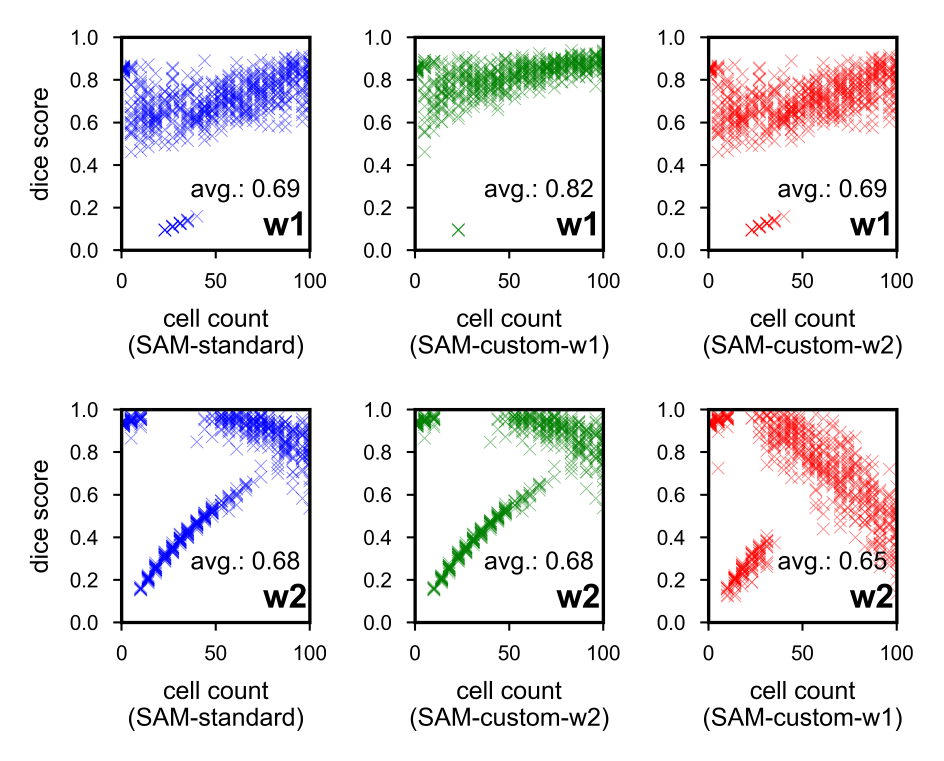}
    \caption{Dice scores, as calculated for segmentation by SAM-standard and SAM-custom of the 600 w1 (top row) and 600 w2 (bottom row) images of the BBBC005 dataset, plotted against cell counts.}
    \label{fig:figure_dice_scores}
\end{figure}

\noindent\textbf{Quantitative evaluation}: In addition to the standard SAM parameters, two more custom parameter settings were finalized, after trial and error (called tuning), in order to obtain segmentations that were better than the standard parameters by minimizing the value of Coefficient of Determination (R-squared or $R^2$) calculated between predicted and actual counts. These trials were performed on BBBC005-S. On these 50 images, the performance of custom parameters that were tuned on w1 (referred to as SAM-custom-w1) when used on the w1 subset ($R^2$ = 0.98) was found to be much better than when standard parameters ($R^2$ = 0.02) were used. However, the performance of a different set of custom parameters that were tuned on w2 (referred to as SAM-custom-w2) when used on the w2 subset ($R^2$ = 0.86) was found to be only marginally better than the standard parameters ($R^2$ = 0.83) despite several attempts at improvement. It appears that the standard parameters were already quite suitable for w2. Please refer to Table \ref{tab:sam_parameters} for the SAM-custom-w1/w2 parameters. Note that the values for finalized SAM-custom-w2 parameters are very close to those of SAM-standard - many other combinations of parameters were tried in order to get to this final set. $R^2$ calculated on cell counts was used as a determinant for these final custom parameters (SAM-custom-w1 and SAM-custom-w2), instead of another metric like dice score, because counting was established as a desired downstream task after segmentation.

Dice scores for each of the 600 w1 and 600 w2 images, as evaluated on the images segmented by SAM, were also obtained. Dice score lies between 0 and 1 and is equal to twice the number of common elements (image pixels, in this case) in both sets divided by the total number of elements present in the two sets. Higher dice score is one indicator of a better segmentation performance in itself, especially in the absence of a known downstream task. Figure \ref{fig:figure_dice_scores} shows the dice scores plotted against cell counts (note that cell counts are available as labels in the BBBC005 dataset) for SAM-standard and SAM-custom. Average dice score for the standard set of parameters was found to be around 0.68 for both w1 and w2 subset (blue plots). As expected, the average dice score for SAM-custom-w1 (green) was better than that for SAM-standard (blue) on the w1 subset, and the average dice score for SAM-custom-w2 (green) was found to be equal to that of standard parameters (blue) on the w2 subset. This was similar and consistent with the $R^2$ values obtained above. For the sake of completeness, SAM-custom-w1 parameters were also used on the w2 dataset (and vice versa) (red plots); dice scores dropped below that of the standard parameters, highlighting the ability of SAM-standard parameters to generalize well. No discernible trend between cell count and dice score was observed.

\begin{figure*}
    \centering
    \includegraphics[width=6.5in]{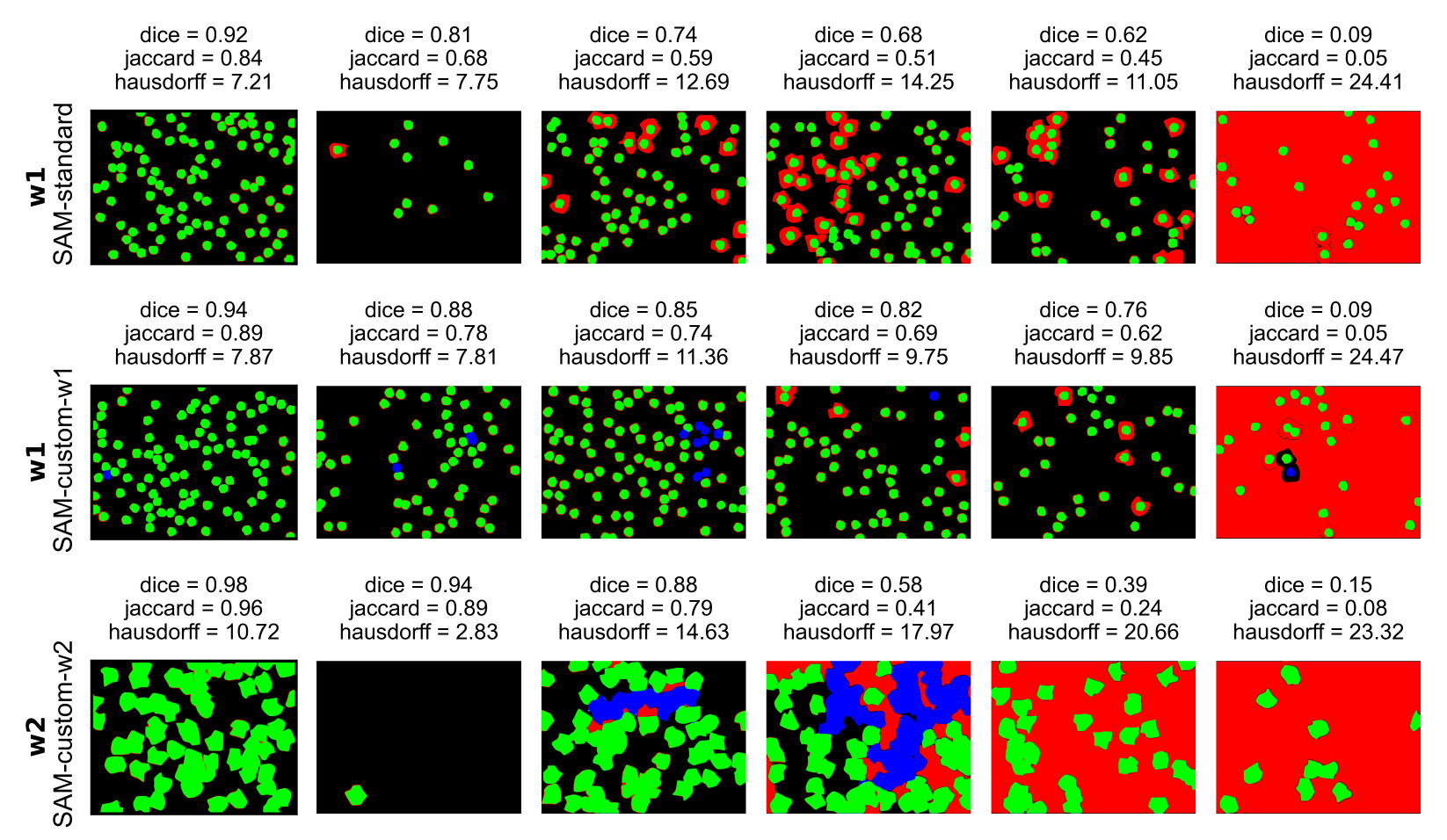}
    \caption{Select \textit{difference images} sorted in decreasing order of dice scores and chosen at equal intervals from the dataset for segmentation of w1 (using SAM-standard and SAM-custom) and w2 (using SAM-custom, which performed very similar to SAM-standard) datasets. F1 scores were the same as dice scores. Jaccard index and Hausdorff distance are also included for reader's reference.}
    \label{fig:figure_diff_images}
\end{figure*}

To further understand the quality of segmentation performed by SAM, difference images were created - if the pixel was present in mask but not in ground truth, it was made red, if present in ground truth but not in mask, it was made blue, and if it was present in both, it was made green. A panel of a few select difference images with a range of dice scores is shown in Figure \ref{fig:figure_diff_images}, along with other popular segmentation scores such as F1 score (equivalent to dice score), Jaccard index (calculated by dividing the size of the intersection of two sets by the size of their union, ranges between 0 and 1 with 0 indicating no overlap and 1 indicating complete overlap), and Hausdorff distance (average of the sum of all minimum distances from all points in one set to all points in another set, lower value being better).

When ChatGPT was asked to perform segmentation, it chose from a variety of computational segmentation techniques. ChatGPT automatically chose a different segmentation algorithm in the event it encountered errors with a given method resulting in performance (as judged by the domain user) ranging from poor to excellent. For the sake of consistency, ChatGPT was eventually instructed to use Otsu's method \cite{otsu_threshold_1979} for segmentation over the 50 image w1-w2 subset. Similar to SAM, dice scores were calculated on the w1-w2 subsets using segmented images downloaded from ChatGPT. ChatGPT scored much higher than SAM in terms of the dice score, averaging 0.96 for w1 and 0.99 for w2 highlighting the accuracy of Otsu's method over SAM for this particular dataset. It shouldn't be overlooked, however, that ChatGPT is capable of trying a variety of segmentation methods computationally, which is deemed of immense utility to domain users.

\begin{figure*}
    \centering
    \includegraphics[width=6.25in]{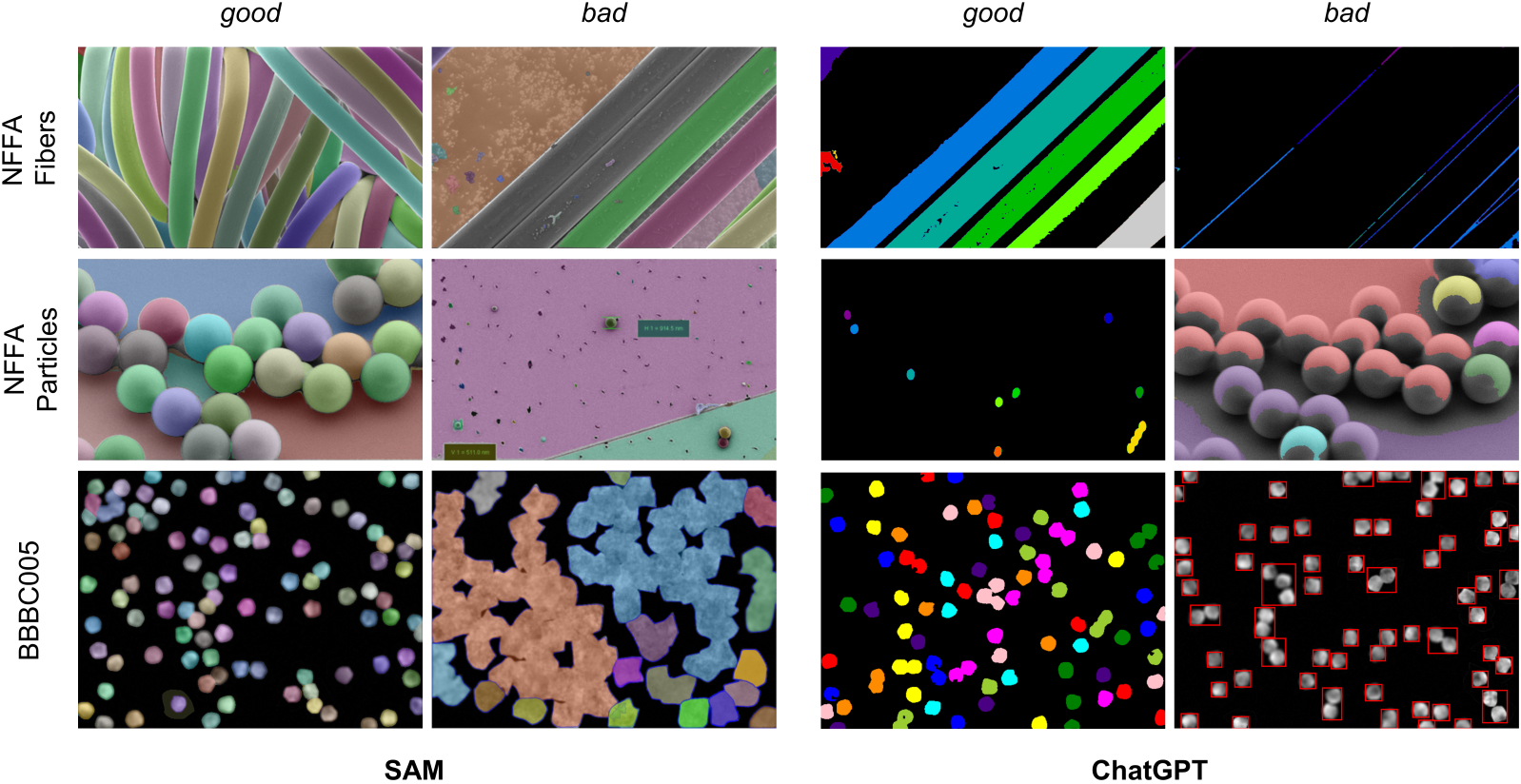}
    \caption{One example each of a good and a bad segmented image output from SAM-standard and ChatGPT on the NFFA-MS (fibers and particles), and the BBC0005-subset. Note that ChatGPT was allowed to use a segmentation algorithm of its own choosing. Ground truths for the BBBC005 are included in the dataset.}
    \label{fig:figure_segmentation_qualitative}
\end{figure*}

\noindent\textbf{Qualitative Evaluation}: Segmentation on the NFFA-MS (containing images from fibers and particles classes) and BBBC005-S, comprising of routinely used SEM images that haven't been thresholded or processed to make segmentation any easier, proved to be quite difficult for both ChatGPT and SAM. In general, SAM (SAM-standard) performed better at extracting objects than ChatGPT. However, in certain instances, depending upon the exact algorithm deployed, ChatGPT was able to perform the segmentation quite well. One example each, of a well segmented and a poorly segmented image has been shown for both ChatGPT and SAM in Figure \ref{fig:figure_segmentation_qualitative}. Furthermore, the authors trust that a more carefully crafted prompt by a domain expert would empower ChatGPT to output impressive results.

\subsection{Counting}
\label{sec:results_counting}

\begin{figure*}
    \centering
    \includegraphics[width=6.5in]{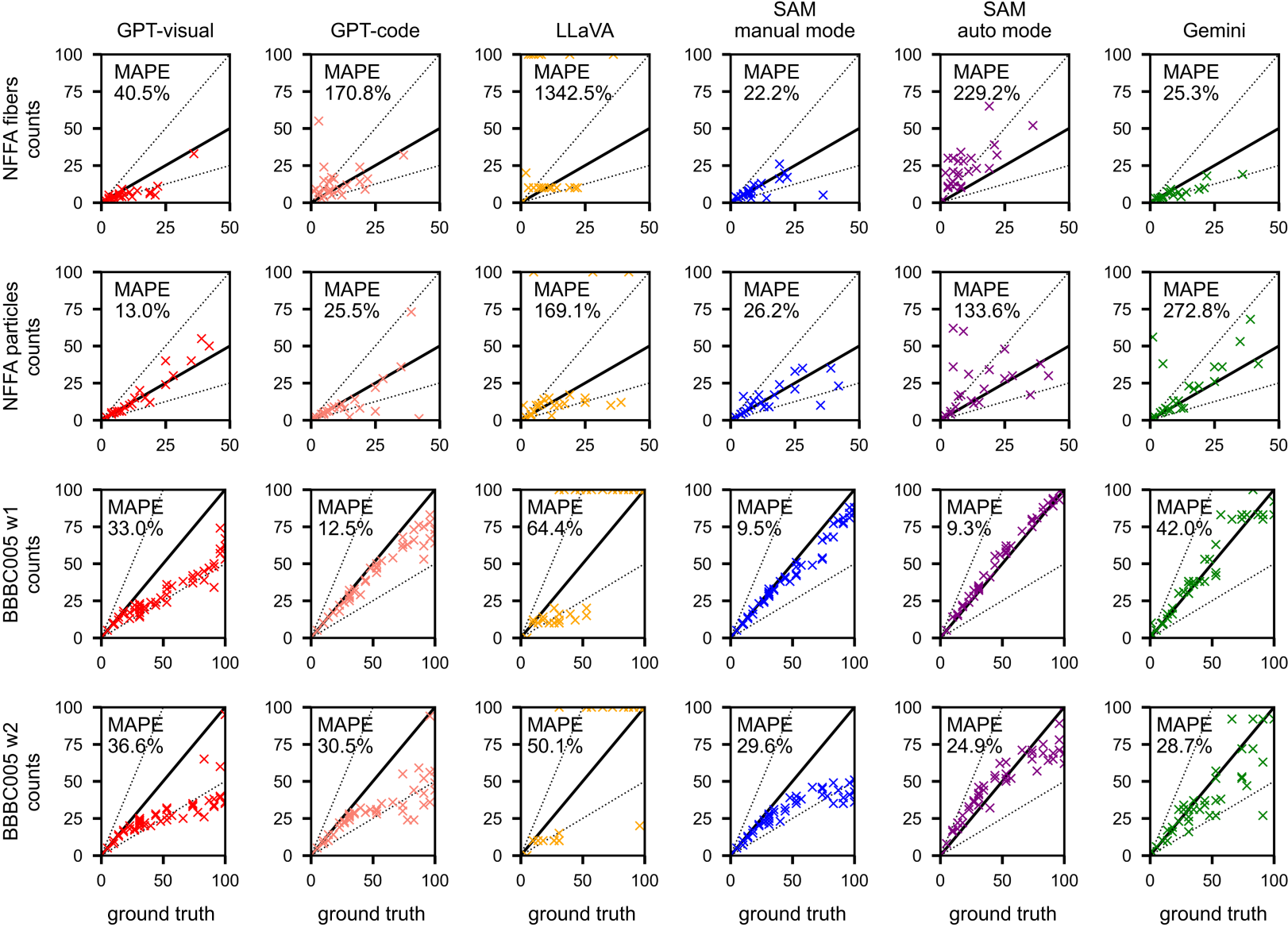}
    \caption{Predicted fiber, particle, and cell counts for sampled data, i.e. NFFA-MS (fibers and particles) and BBBC005-S (w1 and w2), plotted against the actual counts, for ChatGPT, LLaVA, SAM (using standard parameters for NFFA and custom for BBBC005, and used in either manual or automatic counting mode), and Gemini models. Mean absolute percent error (MAPE) is included for quantitative comparison. Ground truth for NFFA datasets refers to the manual count performed by the domain expert.}
    \label{fig:figure_counts_sampled}
\end{figure*}

Results for counting are divided into two sections - one for sampled data, and one for total data. Figure \ref{fig:figure_counts_sampled} shows the fiber, particle, and cell counts as predicted by the VLMs against the ground truths for counts for sampled data. Note that the y-axis has been curtailed to 100 for the sake of clarity which hides some extraneous data. Mean absolute percent error (MAPE), calculated on entire, and not just on displayed data, has been included for quantitative comparison of the closeness of the prediction.

\noindent\textbf{Counting on sampled data.} When using GPT-visual it was observed that ChatGPT underestimated, but was decent at identifying number of fibers (NFFA) and the number of cells (BBBC005), and was excellent at identifying number of particles (NFFA). Note that since the NFFA-MS (particle) dataset has a huge diversity in the type and size of particles, and in the background information, it is deemed quite complex for any standard image processing method. Due to this, the performance of GPT-visual on this dataset is quite noteworthy. When using GPT-code, since the NFFA dataset was more difficult to segment consistently, ChatGPT found it difficult to perform these counts consistently resulting in higher MAPE. For the BBBC005-S (w1), however, ChatGPT was consistently able to segment the cells impressively, which resulted in an accurate cell count, with a MAPE of only 12.5\%. For w2, due to the aggregation of the cells especially for higher counts, both GPT-visual and GPT-code struggled to isolate these cells to get an accurate count.

LLaVA performed the poorest. However, the results on three different datasets provides some insight into the vision-language capabilities of LLaVA. Its performance was better on the easier datasets - BBBC005 is deemed easier than NFFA particles, which in turn is deemed easier than NFFA fibers for visual counting. This indicates that LLaVA is able to make a judgment on the count of the objects in a meaningful way, despite the fact that at times LLaVA showed the tendency to report the count as 1, 5, 10, 100, and so on.

SAM showed a good ability to segment all types of objects in an image in general (as shown before in Figure \ref{fig:figure_segmentation_qualitative}). Segments were produced by SAM-standard for the NFFA subsets and SAM-custom for the BBBC005 subsets. Manual counting, with the exception of BBBC005-S (w2), performed well in-part due to the manual curation of counts and in-part due to SAM's ability to segment out the target artefacts. SAM's segmentation trumped image processing segmentation algorithms employed by GPT-code. Automatic counting of segments produced by SAM-standard performed worse and overestimated counts when compared to manual counting because it segmented all artefacts in the images (not just the targeted artefacts, like fibers only or particles only). This is also evident from the fact that both manual and automatic counting performed almost equally on BBBC005-S (w1) cell counts, because those images contained a single type of isolated artefacts. For w2, SAM-custom-w2 too struggled to identify the boundaries between the agglomerated cells.

Gemini performed very similar to GPT-visual on the NFFA fiber counting, consistently underestimating the number of fibers. On NFFA particles, Gemini's performance fell below all other models with a tendency to overestimate the number. Its performance on BBBC005 was quite impressive though: estimates stayed around the actual counts remarkably well for a model that relied only on visual estimates unlike SAM. Even though the MAPE values were not necessarily lower than those for ChatGPT, perhaps owing to a few outliers, the counts' visible proximity to the y=x line in the figure was distinctively good. Gemini couldn't compete with SAM on the w1 dataset though, but fared fairly well on the w2 dataset showing no signs of underestimation at higher cell counts - a tendency that plagued the other models.

\begin{figure}
    \centering
    \includegraphics[width=3.2in]{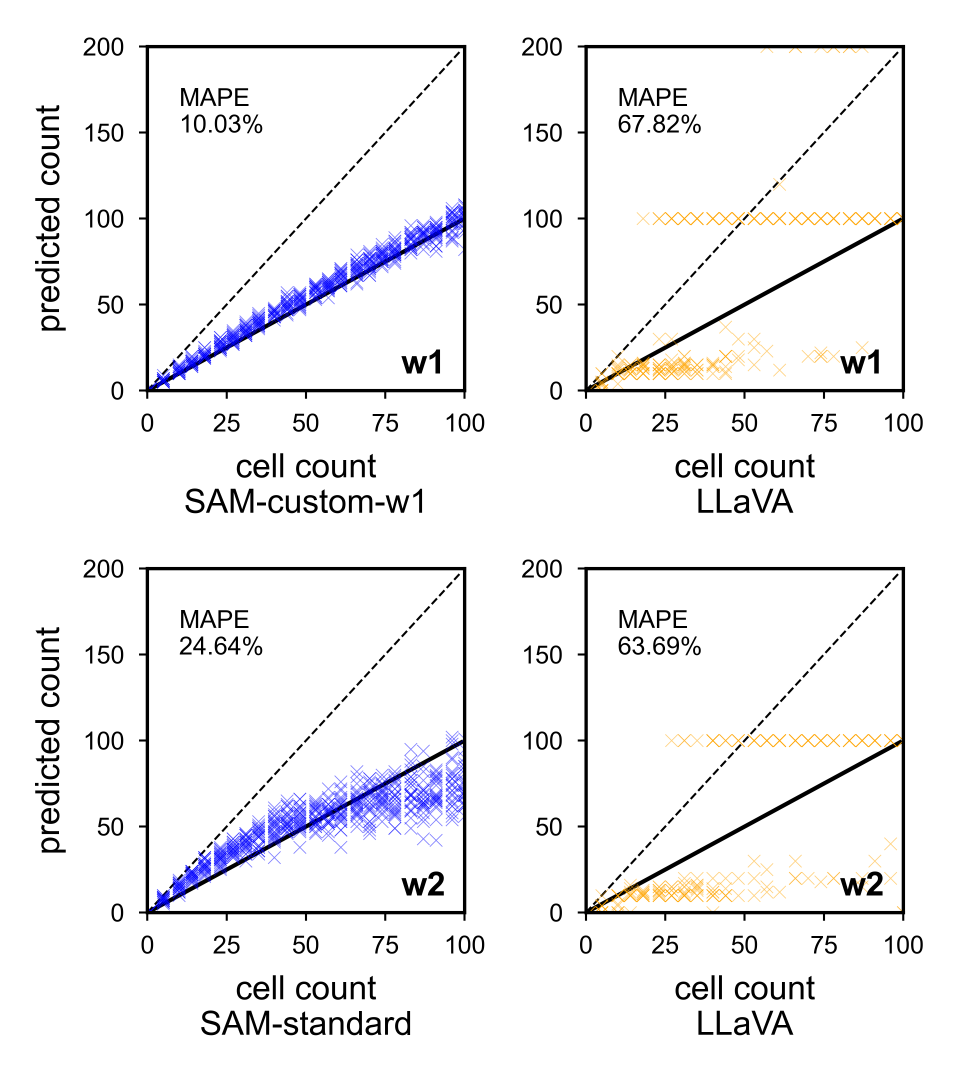}
    \caption{Predicted cell counts for the 1200 image subset of the BBBC005 dataset plotted against the actual counts, for LLaVA and SAM. Counts by SAM on w1 subset are shown for SAM-custom-w1 parameters, while counts on w2 dataset are shown for SAM-standard parameters, both best performing parameters for counting task. Mean absolute percent error (MAPE) is included for quantitative comparison.}
    \label{fig:figure_counts_full}
\end{figure}

 \noindent\textbf{Counting on entire data.} In Figure \ref{fig:figure_counts_full}, the performance of LLaVA and SAM can be observed for the entire 1200 w1 and w2 images from the BBBC005 dataset. Recall that ChatGPT and Gemini could not be used for the entirety of such a large dataset. Results are split for w1 and w2 subsets because some qualitative differences were observed in the results. The performance of LLaVA clearly reveals its tendency to predict counts as 10, 100, 200, and so on. The plots for counting using LLaVA show no qualitative difference between w1 and w2. For SAM, counting was performed using both standard and custom parameters. For w1, the standard parameters resulted in significant overestimation of cell counts (not shown). When the parameters were customized, they resulted in a much improved performance on the w1 subset, resulting in the MAPE going down from 54.25\% (not shown) to 10.03\%. For w2, as also discussed before, the custom parameters did not result in any significant improvements over $R^2$ or dice scores; the MAPE in fact worsened from 24.64\% to 27.31\% (not shown), and therefore, results for the standard parameters have been shown for w2. As the number of cells increased, it became more and more difficult for SAM to segment/segregate cells due to their agglomeration, which, in turn, resulted in a lower estimate at high cell counts. Because there's significantly more agglomeration in w2 compared to w1, this effect was more pronounced in w2.

\subsection{Visual Question Answering (VQA)}
\label{sec:results_vqa}

\begin{figure*}
    \centering
    \includegraphics[width=0.85\linewidth]{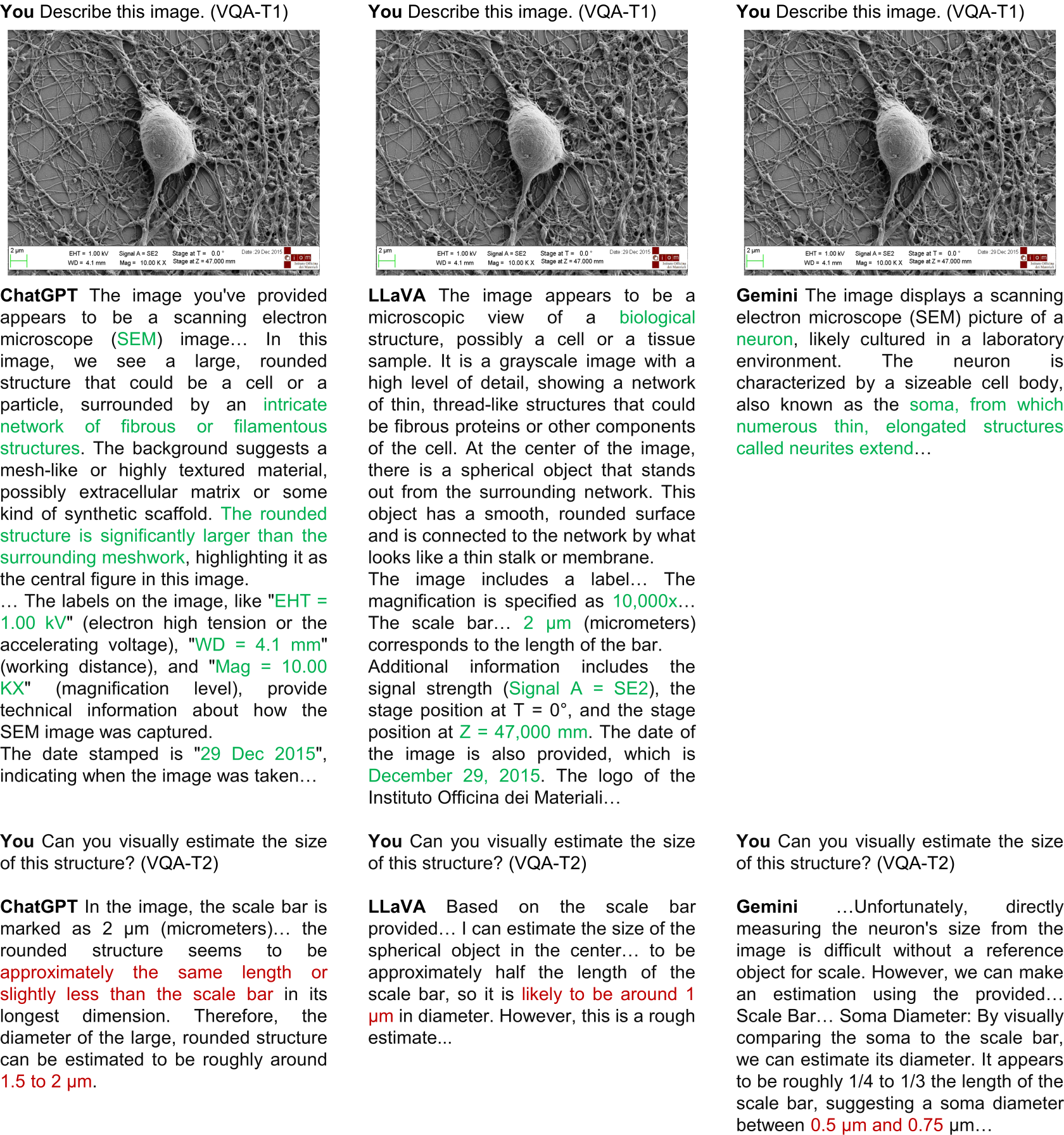}
    \caption{Performance of ChatGPT, LLaVA, and Gemini (left to right) on VQA tasks of describing a biological image and estimating the size of the object of interest. Correct key responses are colored in green, wrong responses in red, and mixed or uncertain in orange. Extraneous information has been replaced with ellipses.}
    \label{fig:figure_vqa_1}
\end{figure*}

\begin{figure*}
    \centering
    \includegraphics[width=0.85\linewidth]{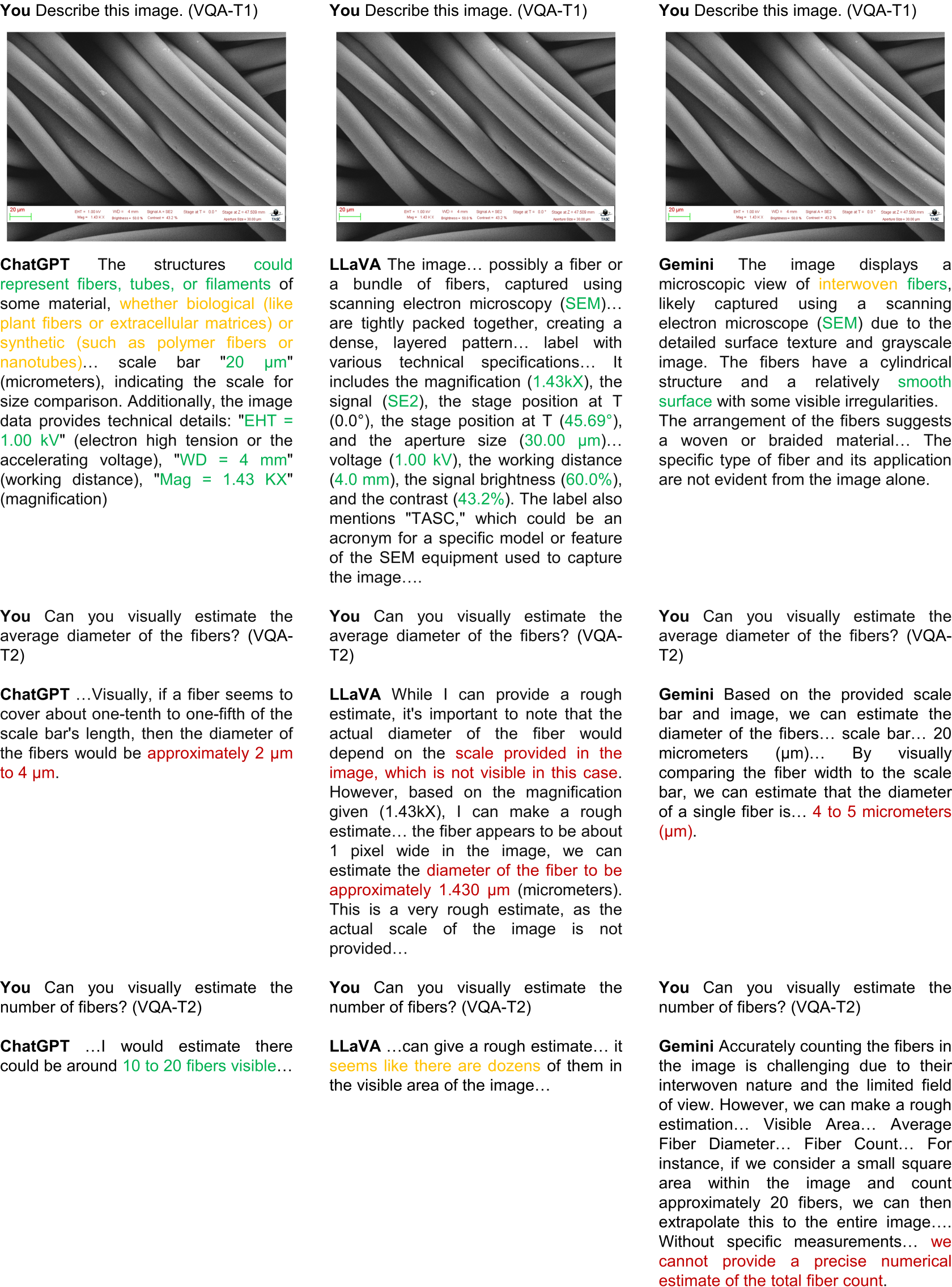}
    \caption{Performance of ChatGPT, LLaVA, and Gemini (left to right) on VQA tasks of describing an image of fibers and estimating the number and average diameter of fibers. Correct key responses are colored in green, wrong responses in red, and mixed or uncertain in orange. Extraneous information has been replaced with ellipses.}
    \label{fig:figure_vqa_2}
\end{figure*}

\begin{figure*}
    \centering
    \includegraphics[width=0.85\linewidth]{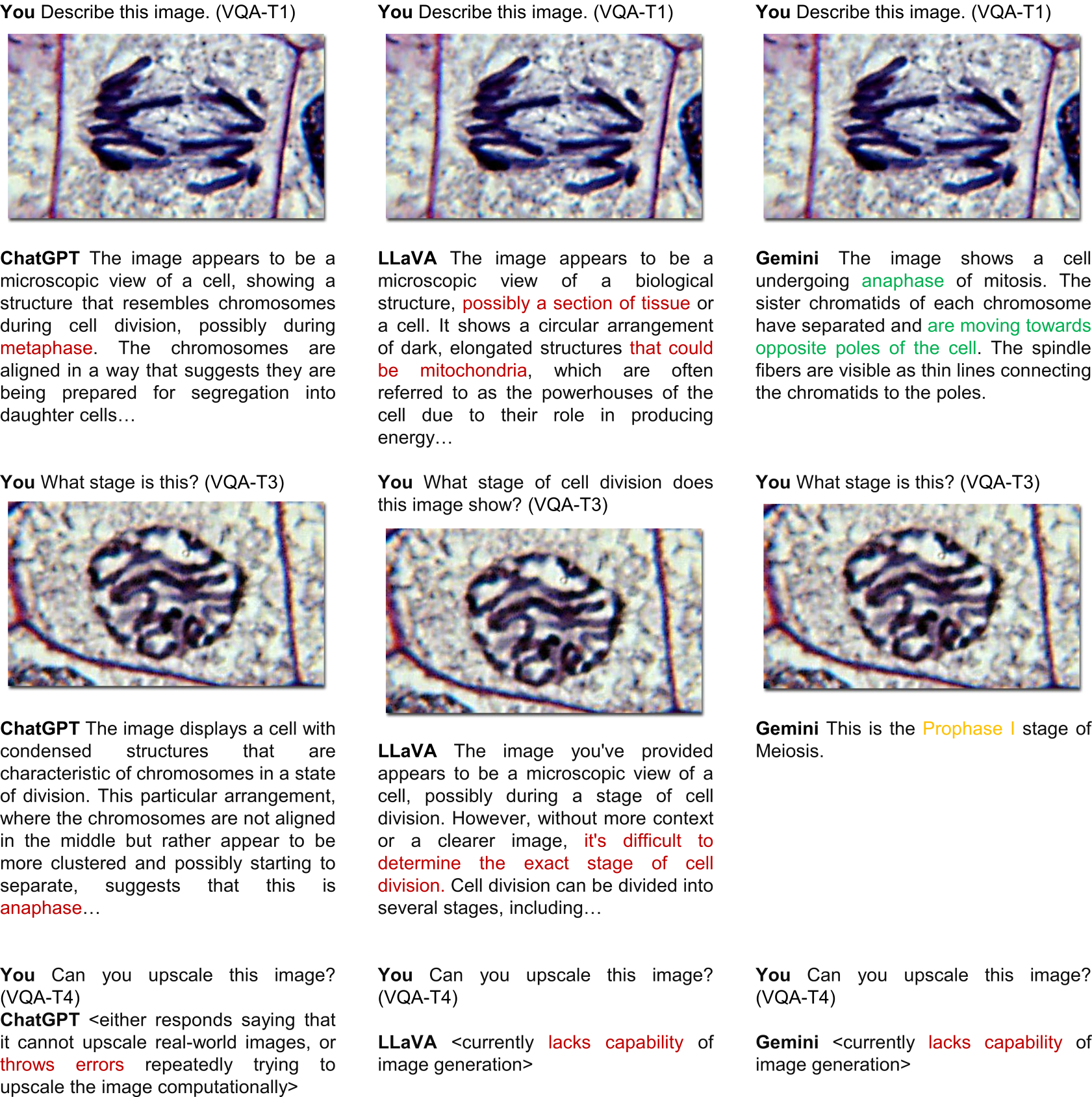}
    \caption{Performance of ChatGPT, LLaVA, and Gemini (left to right) on VQA tasks of describing and identifying the stages of cell division. The ground truth for the top image is `early anaphase' and for the bottom image is `late prophase'. The bottom image was also asked to be upscaled. Correct key responses are colored in green, wrong responses in red, and mixed or uncertain in orange. Extraneous information has been replaced with ellipses.}
    \label{fig:figure_vqa_3}
\end{figure*}

\begin{figure*}
    \centering
    \includegraphics[width=0.85\linewidth]{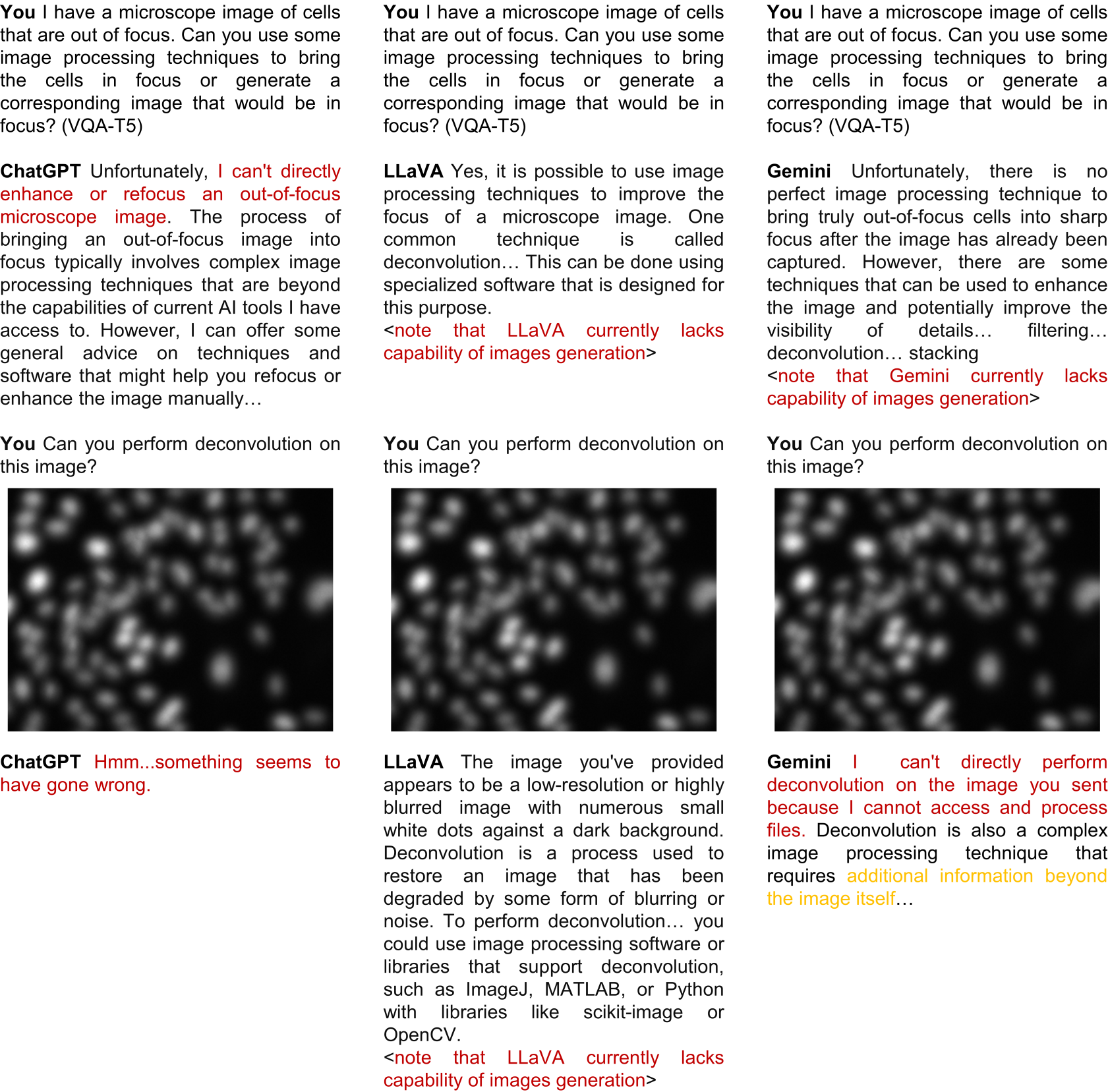}
    \caption{Performance of ChatGPT, LLaVA, and Gemini on VQA tasks of generating an in-focus image from an out-of-focus image. Correct key responses are colored in green, wrong responses in red, and mixed or uncertain in orange. Extraneous information has been replaced with ellipses.}
    \label{fig:figure_vqa_4}
\end{figure*}

The results for VQA tasks performed by ChatGPT, LLaVA, and Gemini are shown in Figures \ref{fig:figure_vqa_1} through \ref{fig:figure_vqa_4}. The models were asked to describe an SEM image of a biological nature (VQA-T1) (see Figure \ref{fig:figure_vqa_1}). All models performed well in identifying the type, nature and the different artefacts present in the image with a satisfactory balance of certainty and uncertainty. All models were accurate at OCR tasks and read all information off of the images perfectly well including the scale bar. In VQA-T2, when asked to estimate the size of the object in the center, the models correctly reasoned the approach of using the scale bar to make such an estimate, but failed at correctly interpreting the size, despite correctly interpreting the size of the scale bar.

Similarly, the models were asked to describe an SEM image of fibers (VQA-T1) and estimate the number and average diameter of the fibers (VQA-T2) (see Figure \ref{fig:figure_vqa_2}). All models were able to identify that the image contained fibers, but seemed uncertain about the nature, origin, or the material composition of the fibers. Like the previous image, they failed at correctly estimating the number or the diameter of the fibers, with the exception of ChatGPT correctly estimating the number.

Subsequently, the models were asked to describe an image of cell-division (VQA-T1) and then identify the stage of cell division on another similar image but showing a different stage (VQA-T3) (see Figure \ref{fig:figure_vqa_3}). In the first image where the models were only asked to describe it, ChatGPT and Gemini succeeded in identifying that the image was indeed of a cellular division but LLaVA failed to understand that. Only Gemini was able to identify the stage close to perfection. When asked to directly identify the stage of the cell division specifically on another similar image, all three models failed to produce the correct answer, but Gemini got close. The models were then asked to upscale the cell division image (VQA-T4). While technically ChatGPT is the only model capable of generating images, it consistently refused to produce an upscaled image despite repeated tries over several days. In VQA-T5 (see Figure \ref{fig:figure_vqa_4}), ChatGPT was asked whether it can generate an in-focus image from an out-of-focus image. Strangely, instead of proceeding to use some known computational technique to generate an improved image, ChatGPT repeatedly refused to perform such an analysis on a scientific image such as this. In the instance where ChatGPT advised to employ deconvolution technique, it was instructed to do so. ChatGPT repeatedly refused to perform deconvolution as well. LLaVA and Gemini too, admitted their inability to perform such operations.

\section{Conclusions}
\label{sec:conclusions}

In this study, we evaluated the performance of ChatGPT, LLaVA, Gemini, and SAM on a series of tasks — classification, segmentation, counting, and visual question answering (VQA) — using a diverse set of microscopy images.

ChatGPT and Gemini excelled in classification tasks, demonstrating robust capabilities despite the relative simplicity of these tasks for a domain expert. LLaVA, in contrast, exhibited weaker performance in classification. All models occasionally confused certain classes, but these instances were logically consistent from both a real-world and scientific perspective, underscoring the models' grasp of visual realism.

SAM distinguished itself in segmentation tasks, adeptly handling scientific images possibly unseen during its pretraining. However, it faced difficulties with artefacts that were clumped together, leading to errors in segmentation and downstream counting. ChatGPT, on the other hand, appeared to employ unpredictable segmentation strategies, which compromised its effectiveness.

In counting tasks, ChatGPT often impressed with its capability to accurately enumerate objects within complex scientific images, even in the presence of various artefacts of non-interest. Gemini's performance was comparable. SAM showed superior performance in scenarios with minimal object ambiguity and diversity present in the image.

The VQA tasks highlighted a significant challenge for the models in comprehending scientific images in depth. They struggled to relate scale bars to object sizes and failed to perform advanced tasks such as identifying stages of cell division, enhancing image resolution, or correcting focus — tasks routinely performed by scientists worldwide.

Overall, while ChatGPT and Gemini showed a commendable understanding of visual features in microscopy images, and SAM was effective in isolating artefacts, the performance of these models does not yet approach the level of a domain expert. The introduction of impurities, defects, overlapping artefacts, and artefact diversity consistently challenged the models, emphasizing the need for further advancements in AI applications for microscopy images.

\section*{Acknowledgements}
This work was supported  in part by National Science Foundation under awards 1946391, the National Institute of General Medical Sciences of National Institutes of Health under award P20GM139768, and the Arkansas Integrative Metabolic Research Center at University of Arkansas.




\end{document}